\documentclass[10pt,twocolumn,letterpaper]{article}

\usepackage{cvpr}
\usepackage{times}
\usepackage{epsfig}
\usepackage{graphicx}
\usepackage{amsmath}
\usepackage{amssymb}
\usepackage{dsfont}
\usepackage{subfigure}
\usepackage{balance}
\usepackage{sidecap}

\graphicspath{{./imgs/}}

\cvprfinalcopy

\def\cvprPaperID{1925} 

\ifcvprfinal\fi
\begin{document}

\title{\vspace{-15pt}Learnt Quasi-Transitive Similarity for Retrieval from Large Collections of Faces}

\author{
Ognjen Arandjelovi\'c\\
University of St Andrews, United Kingdom\\
\small\texttt{ognjen.arandjelovic@gmail.com}\vspace{-5pt}}

\maketitle

\begin{abstract}
  We are interested in identity-based retrieval of face sets from large unlabelled collections acquired in uncontrolled environments. Given a baseline algorithm for measuring the similarity of two face sets, the meta-algorithm introduced in this paper seeks to leverage the structure of the data corpus to make the best use of the available baseline. In particular, we show how partial transitivity of inter-personal similarity can be exploited to improve the retrieval of particularly challenging sets which poorly match the query under the baseline measure. We: (i) describe the use of proxy sets as a means of computing the similarity between two sets, (ii) introduce transitivity meta-features based on the similarity of salient modes of appearance variation between sets, (iii) show how quasi-transitivity can be learnt from such features without any labelling or manual intervention, and (iv) demonstrate the effectiveness of the proposed methodology through experiments on the notoriously challenging YouTube database.
\end{abstract}

\vspace{-2pt}\section{Introduction}\label{s:intro}
The dramatic increase in the capability for large amounts of visual information to be acquired and stored witnessed in the last 10--15 years has effected a profound change on the context in which face recognition algorithms are expected to operate. While the early work on face recognition focused on recognition from a single image using verification and identification protocols on small databases (usually a few dozen people), and at least partly controlled conditions~\cite{MoghPent1997,YangKrieAhuj2002,ZhaoChelPhilRose2004}, more recent efforts have been directed towards video or image set-based recognition~\cite{AranCipo2006e,Aran2014a,HadiPiet2009,LeeHoYangKrie2005}, and large databases acquired in highly uncontrolled environments~\cite{AranCipo2006c,HuanRameBergLear2007,WolfHassMaoz2011}.

Early work on face recognition in the context of large data collections primarily sought to extend existing methods and adapt them for use on low quality images. This includes pose normalization by affine warps \cite{BergBergEdwa+2004} or simplified 3D head models \cite{EverZiss2005}, illumination normalization by filtering~\cite{Aran2009,Aran2012b} and illumination invariance through the use of local gradient-based features~\cite{Aran2012e}. Later work has been increasingly oriented towards challenges associated with learning problems which emerge in large data sets \cite{CaoYingLi2013,WolfHassMaoz2011}. Another popular direction involves the use of text information and natural language processing to extract and associate names with detected faces \cite{EverSiviZiss2009,OzkaDuyg2006}. Concurrently with the research on face recognition in the context of large data collections, there has been much progress in video and set-based recognition\cite{Aran2012,BowyChanFlynChen2006}. Influential contributions include advances in the representation of face sets \cite{ChelSinhPhil2010,SiviEverZiss2005}, and in particular manifold-based representations \cite{LuiBeve2008,WangShanChenGao2008}, illumination models \cite{AranCipo2013}, and similarity measures~\cite{Aran2013,AssaShanAbuq2014,Hu2014,WangShanChenGao2008}.

The broad topic of the present paper is that of face set retrieval and its contribution relates both to the previous work on set-based recognition and the work concerned with recognition in the context of large data collections.
In contrast to most work in the literature our key interest is neither in the representation of face sets nor the associated similarity measures \textit{per se}. Rather, given a baseline algorithm for measuring the similarity of two face sets, our work seeks to leverage the structure of the data at the large scale, that of the entire database, to make the best use of the available baseline. In the sense that our method has as an input both data (face image sets) and an algorithm (the baseline), it can be accurately described as a \emph{meta-algorithm}.

\vspace{-15pt}\paragraph{Problem specification}
Given a query face set our aim is to retrieve from a large database (gallery), sets of the same person. More specifically, we wish to order the gallery sets in decreasing order of confidence that they match the query in identity. Thus the ideal retrieval has all sets of the query person first (`matches') followed by all others ('non-matches'). We assume that the gallery is entirely unlabelled and may contain multiple sets of the same person.

\vspace{-2pt}\section{Learnt transitive similarity}
In this section we introduce the main contribution of the present paper. In particular, we describe a general framework for face retrieval especially well suited for large collections of face images acquired `in the wild' i.e.\ in largely unconstrained imaging conditions, and characterized by highly unbalanced amounts of training data per class (person). We start by motivating the intuition behind our method in the section which follows, and subsequently explain how this intuition can be formalized into a general retrieval framework.

\vspace{-2pt}\subsection{Motivation and the key idea}\label{ss:motiv}
It is useful to consider the motivation behind our idea in the context of related previous work and in particular the recent Matched Background Similarity (MBS) method~\cite{WolfHassMaoz2011}. Wolf \textit{et al.}\ argue that in building a classifier which discriminates the appearance of a specific person and all other people, the focus should be on discriminating between this person and those individuals most similar to him/her; improvements in discrimination against very dissimilar people matter less as these individuals are unlikely to be conflated with the person of interest anyway. Our idea can be seen as complementary and builds upon a similarly simple basic principle. Specifically, we make use of the observation that if person A is alike in appearance to person B, and similarly person B to person C, on average persons A and C are more likely to look alike than two randomly chosen individuals. We term this Quasi-Transitive Similarity; `quasi-' because the stated regularity is a statistical rather than a universal one, as we shall explain shortly.

As stated in our introduction above, the transitivity of similarity in appearance does not hold universally. It is possible that persons A and B are similar by virtue of one set of physical features, and B and C of another. A useful mental picture can be formed by drawing an analogy from statistics (or geometry): random variables (or vectors) A and B, and B and C may be positively correlated (have a positive dot product), yet A and C may be negatively correlated (have a negative dot product) with one another.

Lastly it is worth contrasting our work with that of Yin \textit{et al.}~\cite{YinTangSun2011}. Unlike ours, their method necessitates the localization of face parts, which is problematic and highly likely to fail in severe illuminations, extreme poses, or in poor quality images. Their method also needs to extract estimates of pose and illumination, again very much unlike ours which does not have any of the aforementioned bottlenecks -- all learning is performed directly from data and without the need for an explicit model at a higher semantic level. 

\vspace{-2pt}\subsection{Transitivity meta-features}\label{ss:features}
We have already noted that the observed transitivity of similarity is a statistical rather than a universal phenomenon. In other words, while the similarity of persons A and B, and B and C, on average leads to a greater similarity between A and C, in some instances this will not be the case. This suggests that in addition to  inter-personal similarities A-B and B-C, a richer set of features should be used to infer the similarity A-C. Clearly these features should complement the inter-personal similarities in the sense that jointly they should allow for a better estimate of the similarity A-C than just similarities A-B and B-C, or a direct baseline comparison of A and C (i.e.\ without the use of additional indirect information provided by the relationship of B with A and C).

To motivate the meta-features that we propose in this paper consider the conceptual illustrations shown in Fig~\ref{f:concept}. Solid coloured lines depict the range of appearance variation within face sets. Our aim is to estimate the similarity of the query (green) and the set denoted as `target' (red). The face set marked `proxy' is a database face set of a person similar in appearance to the `target', as assessed by the baseline similarity measure; for example, the proxies of a particular target set can be selected as its nearest $k_p$ sets in the database. The dotted red line represents the range of possible appearance of the `target' person which is not actually present in the `target' face set. For the time being the reader may assume that face sets are represented as sets of actual exemplars and the similarity between two sets is given by the similarity between their most similar members -- we will explain how the ideas introduced herein can be generalized in the next section.

\begin{SCfigure*}
  \centering
  \vspace{-15pt}
  \begin{tabular}{cc}
    \includegraphics[width=0.33\textwidth]{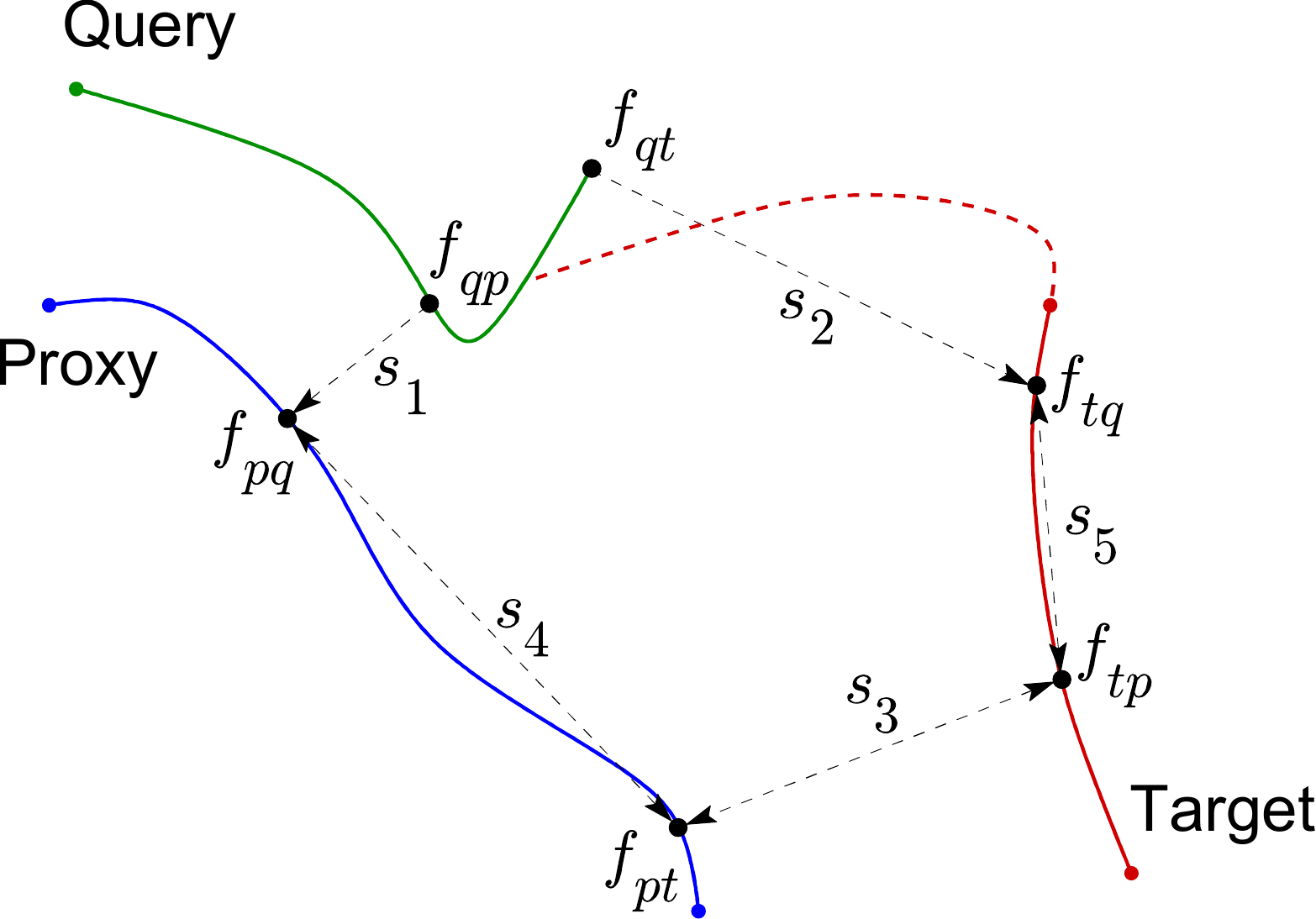}&
    \includegraphics[width=0.38\textwidth]{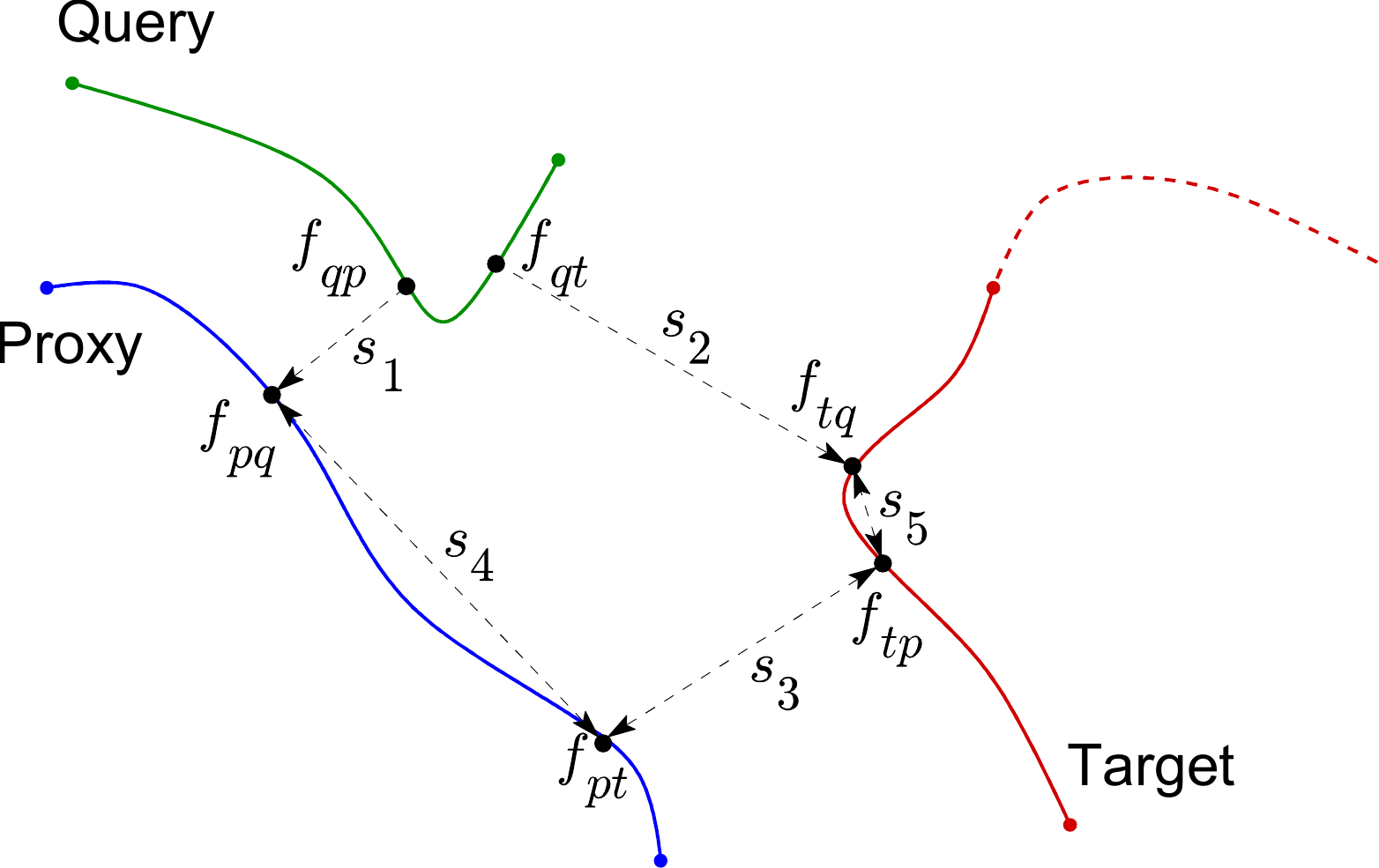}\\
    Query and target: same identity &
    Query and target: different identities\\
    \end{tabular}
    \label{f:concept}
  \caption{Transitivity features extracted using a baseline set comparison: conceptual motivation, using (a) a matching (same identity) query-target set pair, and (b) a non-matching (differing identities) query-target set pair. }
  \label{f:concept}  
  \vspace{-0pt}
\end{SCfigure*}

Both in the case shown in Fig~\ref{f:concept}(a) and that in Fig~\ref{f:concept}(b), the baseline similarity measure tells us that `query' is close to `proxy', and of course `proxy' is close to `target' by design i.e.\ by the former being a proxy in the first place. The difference between the two cases, illustrated conceptually, lies in the similarity of exemplars $f_{tq}$ and $f_{tp}$ i.e.\ the exemplars best matching the query and proxy sets. In particular, the observation that the baseline similarity measure deems the proxy set significantly
more similar than the query to the target on the one hand, while both similarities are explained by similar target exemplars, informs us that the divergence in query and proxy appearances from the target are of different natures. Thus, even if similarities $s_1$, $s_2$, and $s_3$ are the same in Figs~\ref{f:concept}(a) and~\ref{f:concept}(b), the information contained in relationships between $f_{tq}$ and $f_{tp}$, and $f_{pq}$ and $f_{pt}$ tells us that we should infer different query-target similarities in the two cases. Therefore we introduce what we term transitivity meta-features which we use for the said inference. Given a baseline similarity measure and a triplet consisting of query, target, and proxy sets, the corresponding transitivity meta-feature $v(\text{query,target}|\text{proxy})$ comprises five similarities --
$s_1$ (`query' to `proxy' similarity), $s_2$ (`query' to `target' similarity), $s_3$ (`proxy' to `target' similarity), $s_4$ (similarity between the `proxy' exemplar most similar to `query' and the `proxy' exemplar  most similar to `target'), and $s_5$ (similarity between the `target' exemplar most similar to `query' and the `target' exemplar most similar to `proxy'):
{\small\begin{align}
  (\text{query,target}|\text{proxy})=
  \begin{bmatrix}
    s_1 & s_2 & s_3 & s_4 & s_5 \\
  \end{bmatrix}^T
  \label{e:transFeat}
\end{align}}
\vspace{-12pt}

\vspace{-2pt}\subsection{Non-exemplar based representations}
In the preceding discussion we asked the reader to think of appearance variation within each set as being represented using what is probably conceptually the simplest  choice of representation: as a collection of exemplars. In other words, each set was a set of representations of individual faces. This was done for pedagogical reasons and we now show that the proposed framework is in no way reliant on this representation.

In particular, to make the transition of applying the proposed method on the special case in which a face set is represented using a set of directly observed exemplars to the general case in which an arbitrary set representation is employed, we need to explain how the concept of a pair of the most similar exemplars such as those labelled $f_{qp}$ and $f_{pq}$ in Fig~\ref{f:concept}(a), as well as the similarity between them (such as that between $f_{pq}$ and $f_{pt}$), can be generalized. This is not difficult -- all that is required is a slight reframing of the concept. Instead of seeking the nearest pair of specific exemplars, in the general case we are interested in the pair of the most similar modes of variation captured by the representations of two sets (as measured by the baseline similarity measure of course). We illustrate this idea with a few examples.

If the variation within a set is modelled using a linear subspace and the subspace-to-subspace generalization of the distance from feature space (DFFS)~\cite{WangShanChenGao2008} adopted as the (dis)similarity measure between them, the most similar modes of variation between two sets represented using such subspaces are sub-subspaces themselves. These correspond to different exemplars $f_{xy}$ in Fig~\ref{f:concept} and can be compared using the DFFS baseline. If, on the other hand, similarity is measured using the maximum correlation between subspace spans as in \cite{Aran2014}, the most similar modes of variation between two sets are readily extracted as the first pair of the canonical vectors between subspaces~\cite{FukuBachGret2007} and compared using the cosine similarity measure~\cite{NguyBai2010}. For manifold-to-manifold distances such as that of Lee \textit{et al.}~\cite{LeeHoYangKrie2005} the most similar modes of variation are simply the nearest pairs of points on two manifolds, with the similarity of two points on the same manifold readily quantified by the geodesic distance between them.

The same ideas are readily applied to any of a variety of set representations and similarity measures described in the literature.

\vspace{-2pt}\subsection{Learning quasi-transitive similarity}
Given a triplet comprising a query, a target, and a proxy data set, our aim now is to infer the similarity between the query and the target using the corresponding transitivity feature defined in \eqref{e:transFeat}. Without loss of generality, let us quantify inter-set similarity with a real number in the range $[0,1]$, where $0$ signifies the least and $1$ the greatest possible similarity. Then our problem can be stated more formally by saying that we are seeking a mapping $m_\text{qts}$:
{\small\begin{align}
  m_\text{qts}: \mathds{R}^5 \rightarrow [0, 1],
\end{align}}
with the ideal output of $m_\text{qts}(v(\text{query,target}|\text{proxy}))$ being 0 iff the identities in the query and target sets are different, and 1 iff they are the same. Observe that since we are interested in confidence-based ranking of all sets in a database, the codomain of $m_\text{qts}$ is not the set $\{0,1\}$, which would make this a binary classification problem, but rather $[0, 1]$ (a range) which makes it a regression task.

In the types of problem setting in which face recognition is addressed by most of the existing research, obtaining features for training, at least in principle, is simple. Whether it is verification (1-to-1 matching) or identification (1-to-N matching), the database `known' to the algorithm comprises data which is, it is assumed, correctly partitioned by the identity. The retrieval setting adopted in this work is more challenging in this sense and consequently the learning process needs to be approached with more care. In particular, as described in Sec~\ref{s:intro}, we assume that our database is entirely unlabelled and that it may contain multiple sets of the same person. We neither know how many individuals there are in the database nor the number of sets of each individual (which can of course vary person to person). Since for any two database sets we cannot know for certain if they belong to the same or different individuals, an obvious corollary is that in the extraction of transitivity features described by \eqref{e:transFeat} both intra-personal and inter-personal training sets may contain incorrect examples.

\subsubsection{Extraction of transitivity features for training}\label{sss:extractionTraining}
Given that our data is unlabelled i.e.\ that we do not know if the two face sets in the database correspond to the same person or not, we cannot extract training transitivity features in the obvious manner by considering different query, target, and proxy triplets, with the query and the target either matching (producing same identity training data) or not (producing differing identities training data). Instead, we describe how training data, albeit corrupted (this issue is dealt with in the next section), can be collected by considering only pairs of sets, that is, all possible database sets and their proxies. We do this for the two baseline set comparison methods adopted from Wolf \textit{et al.}~\cite{WolfHassMaoz2011}: (i) \textit{maximum maximorum} cosine similarity between sets of exemplars~\cite{NguyBai2010}, and (ii) the maximum correlation between vectors confined to linear subspaces describing within set variability \cite{Aran2014,AranHammCipo2010}.

\vspace{-15pt}\paragraph{Exemplar-based baseline}
Consider a particular database face set (`reference') used for training and one of its proxies. To extract training transitivity features which correspond to same identity query-target comparisons, we select \emph{both} query and target data from the reference set (i.e.\ a single video). In particular, we treat all possible pairs of exemplars in the reference set as possible pairs $f_{qt}$ and $f_{tq}$. Indeed, for specific choices of possible query and reference sets, any two appearances may present themselves as the nearest exemplars in them. The second element $s_2$ in the transitivity feature is then simply given by the similarity between the two exemplars. On the other hand the similarity $s_1$ between the query and the proxy is given by the similarity between the unitary set consisting of the reference set exemplar treated as $f_{qt}$ and the proxy set. The nearest proxy exemplar to $f_{qt}$ is of course $f_{pq}$. The similarity $s_3$ is simply computed as the similarity between the reference set and the proxy, which also gives us exemplars $f_{pt}$ and $f_{tp}$, and allows for a straightforward computation of $s_4$ (as the similarity between $f_{pq}$ and $f_{pt}$) and $s_5$ (as the similarity between $f_{tq}$ and $f_{tp}$). A single pair of reference and proxy sets thus gives us $n_r (n_r-1)$ `positive' training transitivity features, where $n_r$ is the number of faces in the reference set.

The extraction of training transitivity features which correspond to differing identities query-target comparisons is similar. Now we iterate through all exemplar pairs of the proxy set, taking each pair as $f_{qt}$ and $f_{pq}$ in turn. The closest target exemplar to $f_{qt}$ becomes $f_{tq}$, while $f_{pt}$ and $f_{tp}$ are determined as before, allowing for all transitivity feature entries (exemplar similarities) to be computed as in the case of same identity query-target training data extraction. A single pair of reference and proxy sets thus gives us $n_p (n_p-1)$ `negative' training transitivity features, where $n_p$ is the number of faces in the proxy set.

It is important to observe that the set of `negative' training transitivity features extracted in the described manner may be corrupt. This is an inherent consequence of the problem setting -- since the database is entirely unlabelled we cannot know if the identities of the people in the reference and proxy set are actually different. The proposed process of training the regressor, described in Sec~\ref{sss:eSV}, takes this into account. Nevertheless, the amount of improvement achieved with the proposed method over its baseline is tied to the proportion of `negative' training data which is incorrect -- the improvement inevitably decreases as this proportion  is increased. However, if this is so, i.e.\ if a great proportion of proxies of sets in the database actually represent the same identity as the sets they are proxies to, this by design means that the baseline comparison is very good to start with so no significant improvement can be reasonably expected. Thus, our method is particularly attractive in challenging conditions in which the baseline classifier does not perform well.

\vspace{-15pt}\paragraph{Subspace-based maximum correlation baseline}
The extraction of training data for this representation is somewhat simpler than in the previous case. We again extract transitivity feature training data using only face set pairs (rather than triplets) which are now represented by linear subspaces. To extract training transitivity features which correspond to same identity query-target comparisons, we iterate through all reference set exemplars as $f_{qt}$ and obtain $f_{tq}$ and $f_{pq}$ by projecting them to respectively the reference and proxy subspaces. Vectors $f_{pt}$ and $f_{tp}$ are readily obtained using the baseline set comparison as the principal vectors of the subspaces corresponding to reference and proxy subspaces. A single pair of reference and proxy sets thus gives us $n_r$ `positive' training transitivity features.

The extraction of training transitivity features which correspond to differing identities query-target comparisons proceeds in exactly the same manner, with the difference that it is proxy set exemplars that are iterated through as $f_{qt}$ (as before also taken to be $f_{qp}$).
A single pair of reference and proxy sets gives us $n_r$ `positive' training transitivity features, where $n_r$ is the number of faces in the reference set, and $n_p$ `negative' training transitivity features, where $n_p$ is the number of faces in the proxy set. A single pair of reference and proxy sets thus gives us $n_p$ `negative' training transitivity features.  The same remarks as before regarding the corruption of the `negative' training set hold here too.

\vspace{-15pt}\paragraph{Closing note} In Sec~\ref{ss:motiv} we remarked that the basic idea behind the proposed method can be seen as complementary to that of MBS~\cite{WolfHassMaoz2011}. However when the proposed training scheme is considered it can be seen to contain both conceptually similar elements \emph{and} complementary elements to MBS. In particular, since the negative training set of quasi-transitivity features is extracted by considering elements of the proxy set as the query, our method learns to discriminate precisely between a person and those individuals most similar to him/her (as in MBS), while exploiting the quasi-transitivity of similarity (complementary to MBS).

\subsubsection{Training the predictor}\label{sss:eSV}
We adopt the use of the $\epsilon$ support vector ($\epsilon$-SV) regression~\cite{Vapn1995}. For comprehensive detail of this regression technique the reader is referred to the original work by Vapnik; here we present a brief summary of the ideas relevant to the proposed method. Given training data $\{ (x_1,y_1),\ldots,(x_l,y_l) \} \subset \mathcal{F} \times \mathbb{R}$, where $\mathcal{F}$ is the input space (in our case this is $\mathds{R}^5$), $\epsilon$-SVR aims to find a function $h(x)$ which deviates at most $\epsilon$ from its targets $y$. As in other SV-based methods, an implicit mapping of input data $x \rightarrow \Phi(x)$ is performed by employing a Mercer-admissible kernel~\cite{Merc1909} $k(x_i,x_j)$ which allows for the dot products between mapped data to be computed in the input space: $\Phi(x_i)\cdot \Phi(x_j)=k(x_i,x_j)$. The function $h(x)$ of the form
{\small\begin{align}
  h(x)=\sum_{i=1}^l(\alpha_i-\alpha_i^*)k(x_i,x)+b \\[-18pt] \notag
\end{align}}
is then learnt by minimizing
{\small\begin{align}
  \sum_{i=1}^l \sum_{j=1}^l &(\alpha_i-\alpha_i^*)(\alpha_j-\alpha_j^*)k(x_i,x_j) \notag \\[-11pt]
  +&\epsilon\sum_{i=1}^l(\alpha_i+\alpha_i^*)-\sum_{i=1}^l y_i(\alpha_i-\alpha_i^*) \\[-20pt] \notag
\end{align}}
subject to the constraints $\sum_{i=1}^l(\alpha_i-\alpha_i^*)=0$ and $\alpha_i, \alpha_i^* \in [0,c]$. The parameter $c$ can be seen as penalizing prediction errors greater than $\epsilon$ i.e.\ as balancing the trade-off between the smoothness of $h(x)$ and the amount of data predicted with an error greater than $\epsilon$.

The nature of $\epsilon$-SV regression is particularly well suited to the problem at hand. 
Specifically, we train the regressor using the value of 1 as the target for same identity transitivity features, and 0 for different identities, allowing for a large prediction error margin of $\epsilon=0.4$ but severely penalizing greater errors with $c=1000$. The large penalty $C$ ensures that it is the outliers in the form of the wrongly labelled training data that define the boundary between the penalized and non-penalized regions of the high-dimensional space, while the wide margin $\epsilon=0.4$ ensures that the correctly labelled bulk of the training corpus is pushed away from the boundary towards the desired extreme values of 0 and 1. We used the radial basis function kernel $k(x_i,x_j)=\exp\{-0.2\|x_i-x_j\|^2\}$.

\subsubsection{Retrieval}
Given a query data set we compute its similarity with a target database set by computing the regression-based estimate $m_\text{qts}(v(\text{query,target}|\text{proxy}))$ using each of target's $k_p$ proxies, and taking the maximum of these and the baseline similarity between the query and the target. Database sets are then ordered by decreasing similarity with respect to the query.

\begin{figure}[htb]
  \centering
  \vspace{-8pt}
  \includegraphics[width=0.4\textwidth]{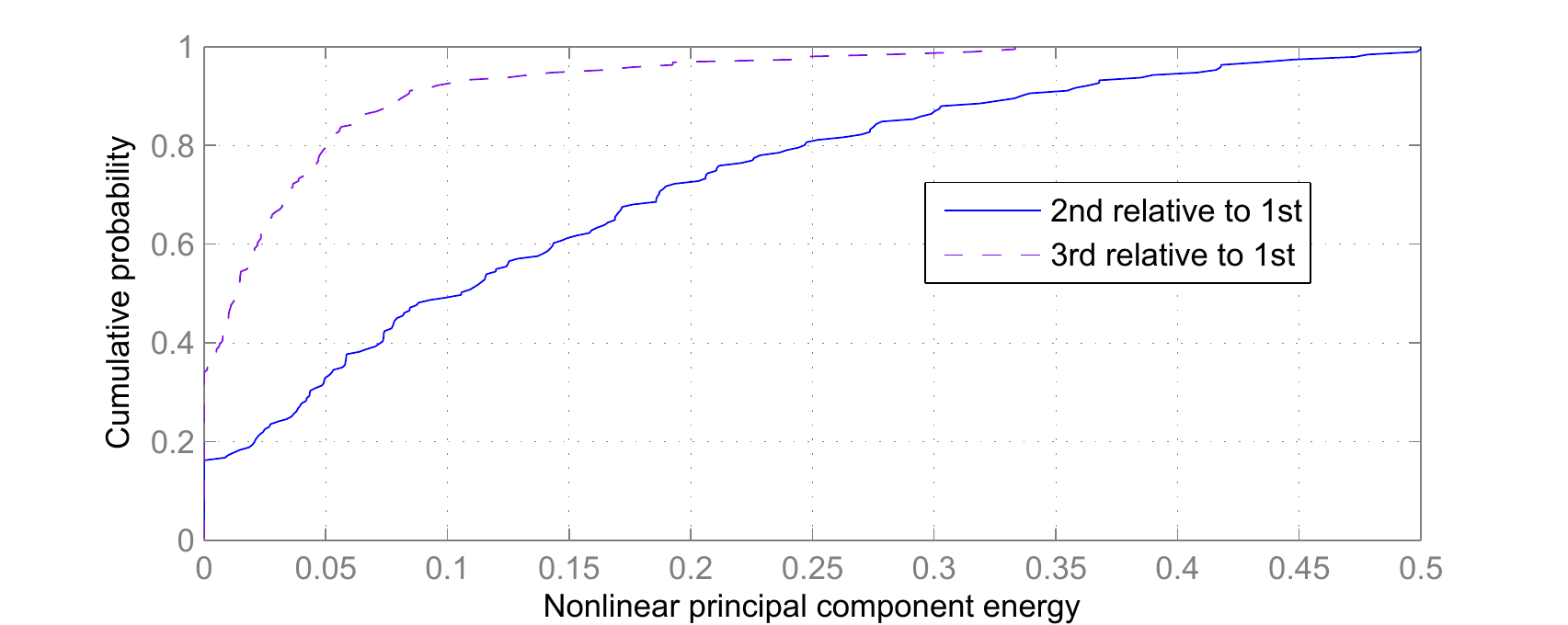}
  \caption{The cumulative distribution function (CDF) of the data energy contained in the 2nd and 3rd nonlinear kernel PCA components relative to the energy of the 1st component, across sets in the YouTube Faces Database. The variation within sets is strongly dominated by the 1st nonlinear principal component.}
  \label{f:kpcaEnergy}
\end{figure}

\begin{figure}[htb]
  \centering
  \vspace{-10pt}
  \includegraphics[width=0.35\textwidth]{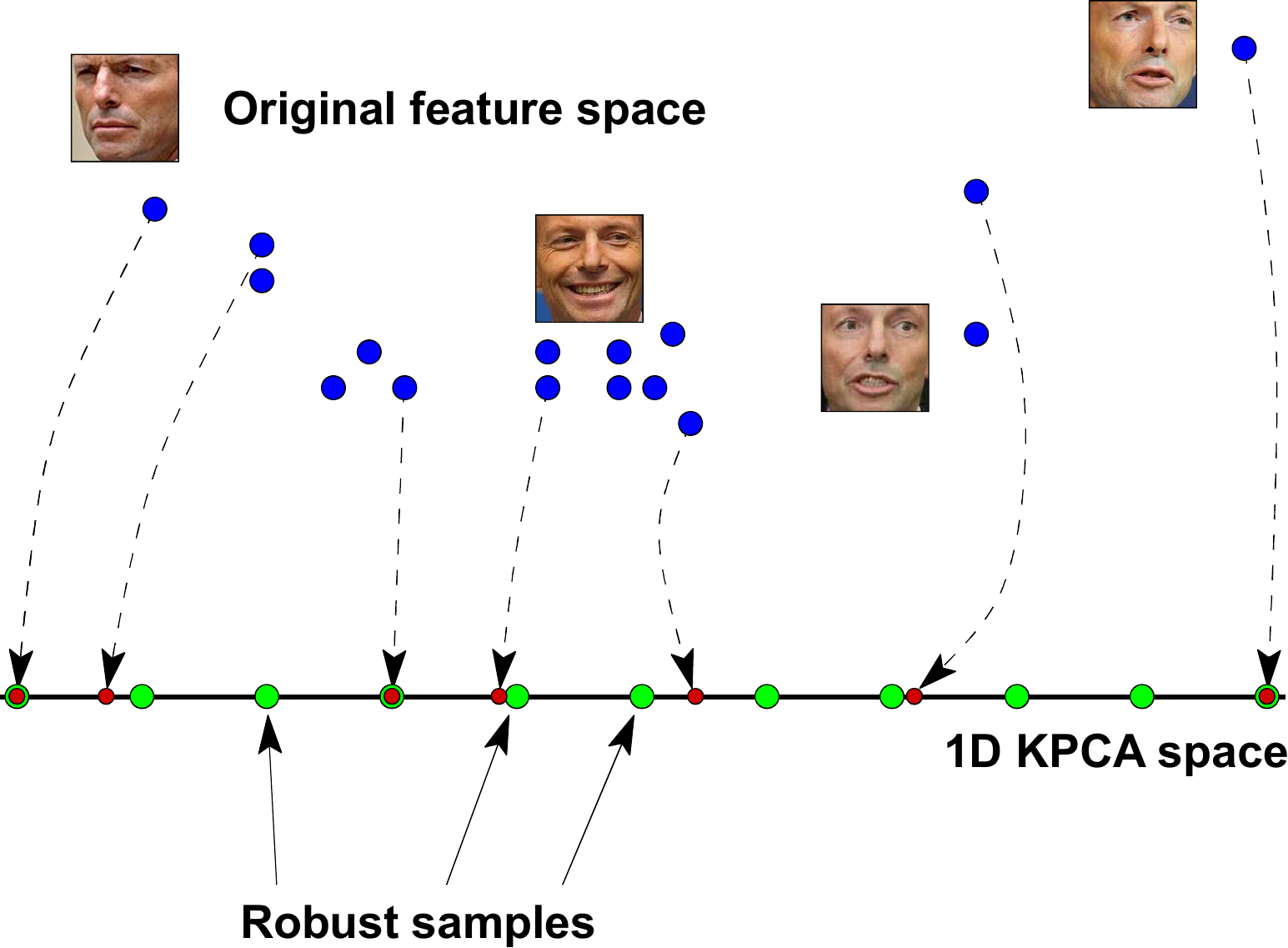}
  \caption{Conceptual illustration of our robust sample selection: (i) original exemplars are projected onto their 1st kernel principal component, (ii) uniform sampling between the extreme projections is performed in the 1D kernel space, and (iii) the obtained samples are re-projected into the original space (step not shown). }
  \label{f:kpca}
  \vspace{-10pt}
\end{figure}

\begin{figure}[htb]
  \centering
  \includegraphics[width=0.4\textwidth]{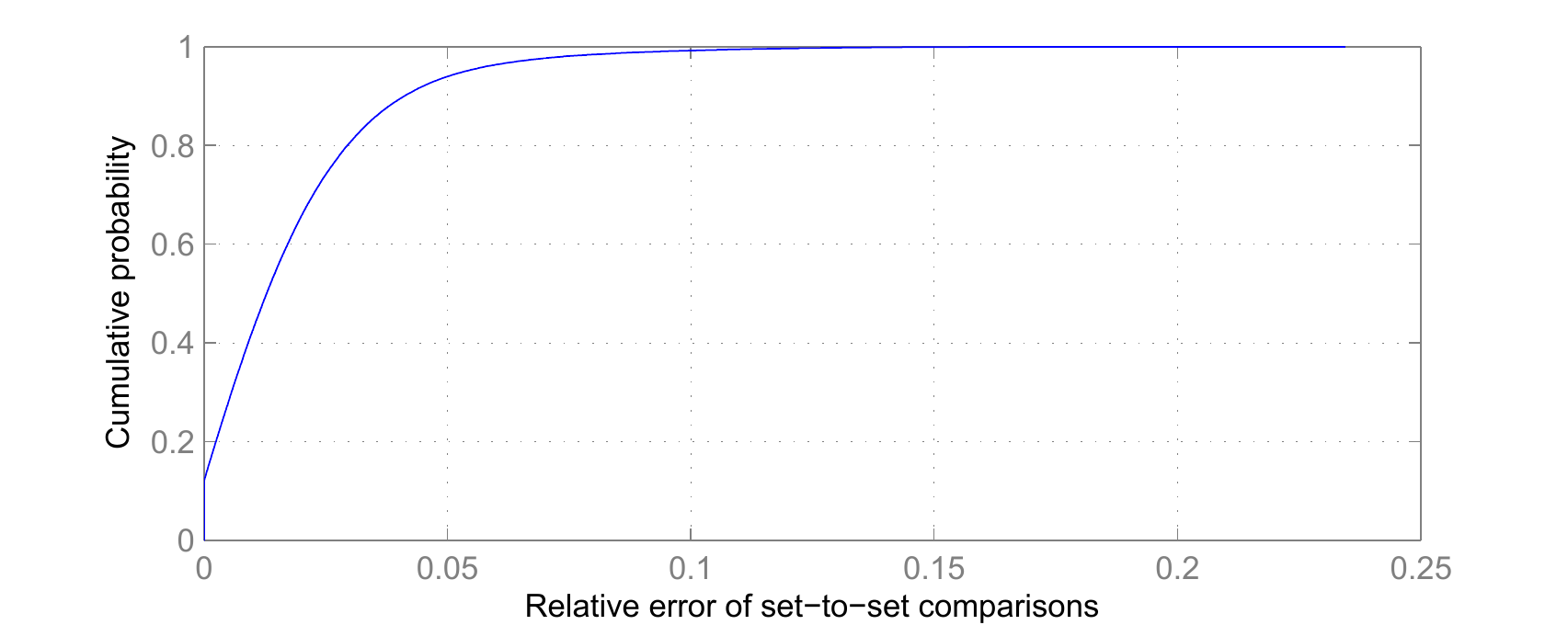}
  \caption{CDF of the error introduced by our robust sample selection (10 samples were used) in the exemplar-based set method. }
  \label{f:samplingError}
  \vspace{-15pt}
\end{figure}

\vspace{-2pt}\section{Evaluation}\label{s:eval}
In this section we report our evaluation of the proposed method and discuss our findings. We start by describing the data set on which the evaluation was performed, consider the measures used to assess performance, summarize the evaluated baseline set representations, distances and their derivatives, and finally present and comment on the results. 

\vspace{-2pt}\subsection{Evaluation data}\label{ss:data}
For evaluation we adopted the YouTube Faces Database~\cite{WolfHassMaoz2011} which contains sets of faces extracted from YouTube videos. There are two key reasons which motivated this choice. Firstly, the manner in which this data set was collected and the nature of its contents are representative of the conditions which the present work targets. In particular, the total amount of data is large (3425 videos/sets of 1595 individuals, with the average set size of approximately 181.3 faces or equivalently 620,953 faces in total), it was extracted from videos acquired in unconstrained conditions in which large changes in illumination, pose, and facial expressions are present, and the distribution of data is heterogeneous both with respect to the set sizes (48--6,070) as well as the number of sets (1--6) for each person in the database. The second reason lies in the reproducibility of results and the ease of comparison with alternatives in the literature -- the database has been widely adopted as a standard benchmark and a number of standard face representations are provided ready for use. Full detail can be found in the original publication~\cite{WolfHassMaoz2011}.

\vspace{-2pt}\subsection{Performance evaluation}
As the cornerstone measure of retrieval performance we adopt the average normalized rank (ANR) \cite{DeseKeysNey2004,SaltMcGi1983}. In brief, ANR treats each retrieved datum as either matching or not matching the query and computes the average rank of the former group, normalized to the range $[0,1]$, with the ANR value of 0 corresponding to the best possible performance (all matching data retrieved before any non-matching) and 1 the worst (all non-matching data retrieved before any matching). Formally:
{\small\begin{align}
   ANR(n, \{r_1,\ldots,r_c\}) = \frac{\sum_{i=1}^c r_i - m}{M-m}\\[-18pt] \notag
\end{align}}
where $n$ is the database size, $\{r_1,\ldots,r_c\}$ the set of retrieval ranks corresponding to the data of interest (i.e.\ data matching
the query), and $m$ and $M$ respectively the minimum and maximum possible values of the sum of $r_1,\ldots,r_c$:
{\small\begin{align}
   m &= \sum_{i=1}^c i = \frac{c\times(c+1)}{2}\\[-5pt]
   M &= \sum_{i=n+1-c}^n i = 
     c\times\frac{2n-c+1}{2}\\[-18pt] \notag
\end{align}}
In comparison with other common performance measures, such as the receiver operating characteristic (ROC) curve~\cite{Fawc2006}, commonly used in verification and identification problems~\cite{Aran2014b}, the average normalized rank more directly captures the ultimate aim of a retrieval algorithm.

\vspace{-2pt}\subsection{Methods}\label{ss:methods}
Motivated by the results reported by Wolf \textit{et al.} which demonstrate its superiority over a number of alternatives, we adopt the standard local binary pattern (LBP) representation of individual faces~\cite{HeikPietSchm2009}. Using LBP we consider two baseline set representations: (i) a set of LBP exemplars, and (ii) a linear LBP subspace, both of which were also evaluated by Wolf \textit{et al.} The former simply stores all face exemplars (i.e.\ the corresponding LBP vectors), while the latter uses PCA to represent the main modes of the observed exemplar variation; previous work suggests that for individual face sets 6-dimensional subspaces produce good results so this is the dimensionality we adopt too.

We adopt two baseline set similarity measures, again motivated by the reports of their good performance in the existing literature. The first of these is the \textit{maximum maximorum} (`max-max') cosine similarity between sets of exemplars $\max_{f_1 \in S_1,f_2 \in S_2} f_1^T f_2 / \|f_1\| /\|f_2\| $ which in the experiments of Wolf \textit{et al.}~\cite{WolfHassMaoz2011} outperformed a number of alternatives including by a large margin the pyramid match kernel of Graumanand and Darrell~\cite{GrauDarr2005} and the locality-constrained linear coding (LLC) of Wang \textit{et al.}~\cite{WangYangYuLv+2010}. The second baseline comparison which we adopt for the comparison of sets represented as linear subspaces is the algebraic method based on the maximum correlation between pairs of vectors lying in two subspaces. This method too performed well in past experiments~\cite{WolfHassMaoz2011,Aran2014}. Thus in summary, our two baseline methods are:
\vspace{-5pt}\begin{itemize}
  \item LBP + \textit{maximum maximorum} set similarity, and\\[-17pt]
  \item LBP + maximum correlation between subspaces.
\end{itemize}\vspace{-2pt}
These are used to establish reference performance. They are then employed in the context of several different ways of applying our idea of quasi-transitivity:
\vspace{-4pt}\begin{itemize}
  \item Simple arithmetic mean-based quasi-transitivity,\\[-17pt]
  \item Simple geometric mean-based quasi-transitivity,\\[-17pt]
  \item Simple quadratic mean-based quasi-transitivity, and\\[-17pt]
  \item Proposed learnt quasi-transitivity (L-QTS).
\end{itemize}\vspace{-2pt}
The first three methods in the list are simple combination rules. In the first of these, the arithmetic mean-based quasi-transitivity, two set similarity of
dissimilarity measures $\rho_{QP}$ (query-proxy) and $\rho_{PT}$ (proxy-target) are combined by computing their arithmetic mean i.e.\ $0.5\times (\rho_{QP}+\rho_{PT})$. Similarly, in the geometric and quadratic mean-based methods quasi-transitivity is attempted by computing respectively $\sqrt{\rho_{QP} \times \rho_{PT}}$ and $\sqrt{0.5{\rho_{QP}}^2+0.5{\rho_{PT}}^2}$. The proposed learnt quasi-transitivity (applied atop of both baseline methods) was evaluated using different numbers of proxy sets (1--10) and as detailed in Sec~\ref{sss:eSV}, $\epsilon$-SV regression was learnt using the parameter values $\epsilon=0.4$ and $c=1000$.

\begin{figure*}[htbp]
  \centering
  \vspace{-15pt}
  \subfigure[Exemplar b/line, var.\ methods]{\includegraphics[width=0.235\textwidth]{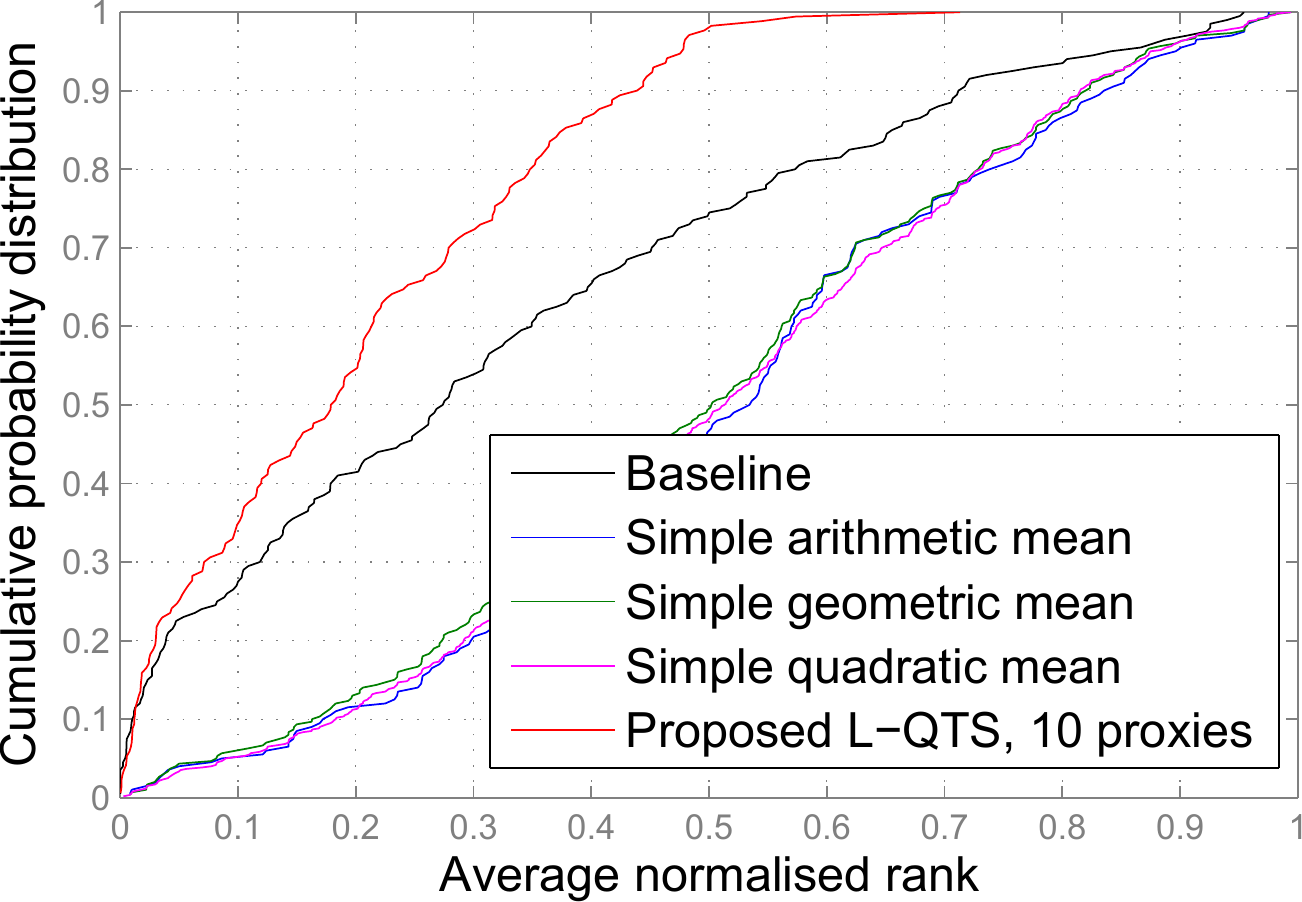}\label{f:anrSamp}}~~~
  \subfigure[Exemplar b/line, proposed]{\includegraphics[width=0.235\textwidth]{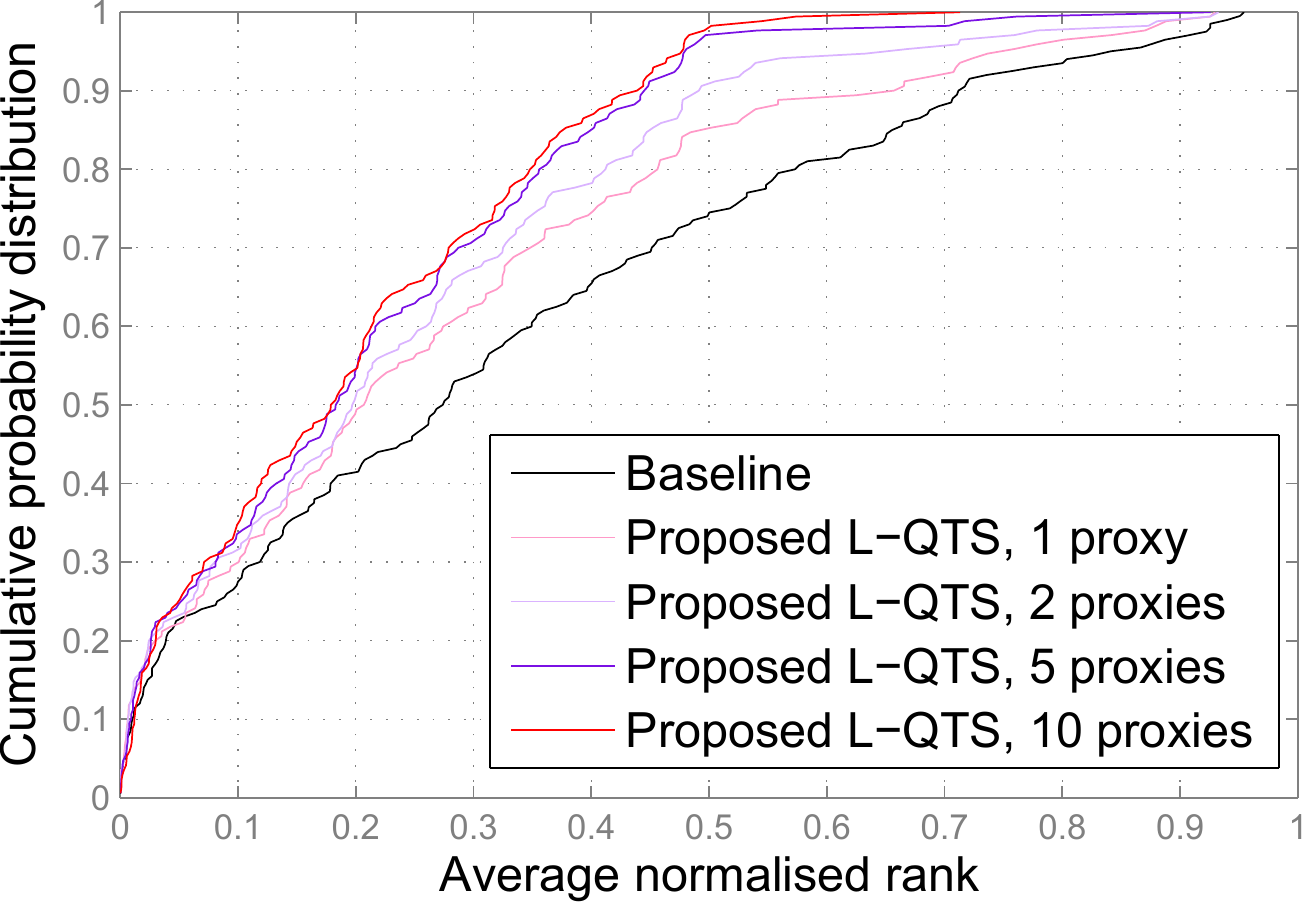}\label{f:anrSampNN}}~~~
  \subfigure[Subspace b/line, var.\ methods]{\includegraphics[width=0.235\textwidth]{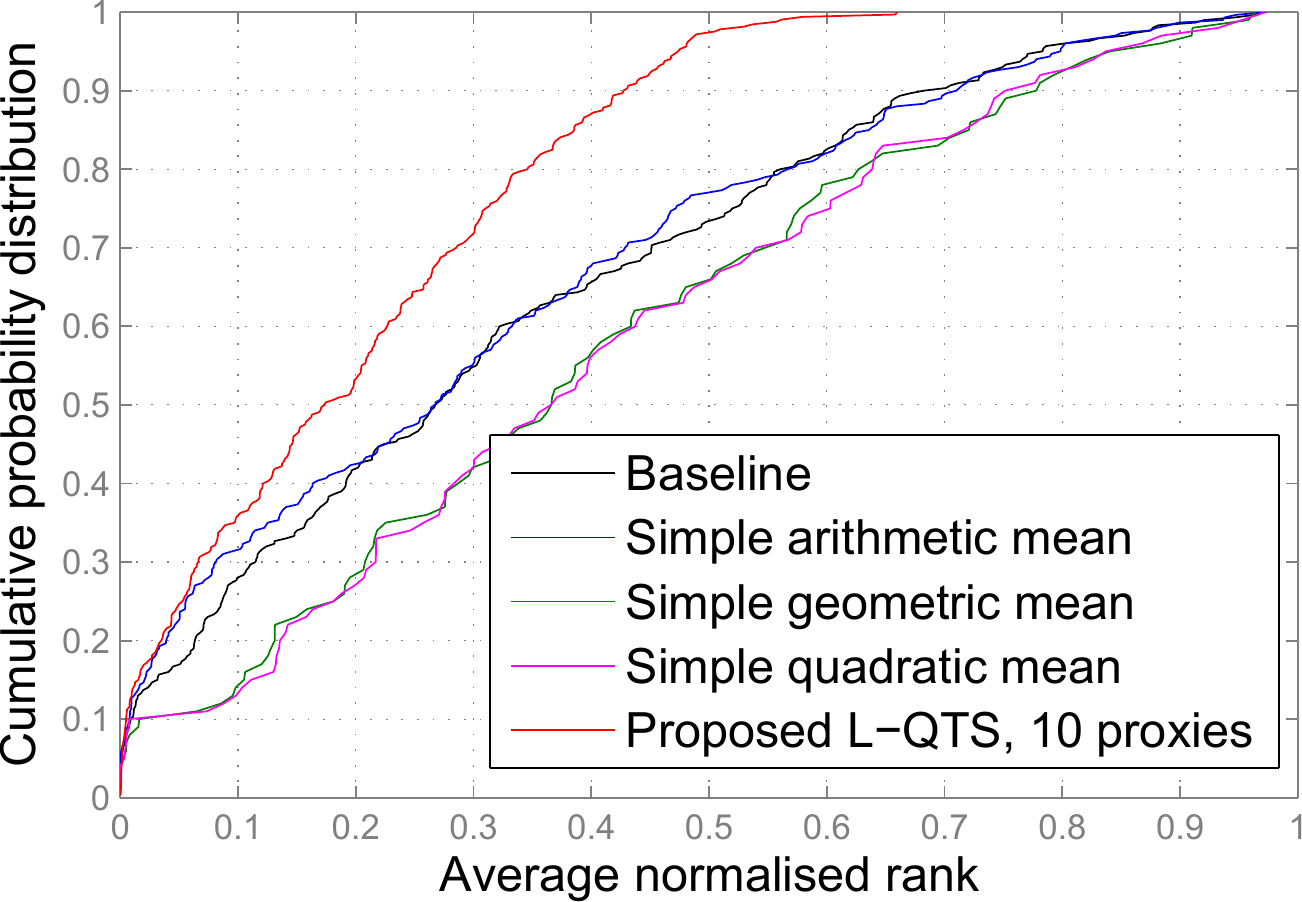}\label{f:anrCCA}}~~~
  \subfigure[Subspace b/line, proposed]{\includegraphics[width=0.235\textwidth]{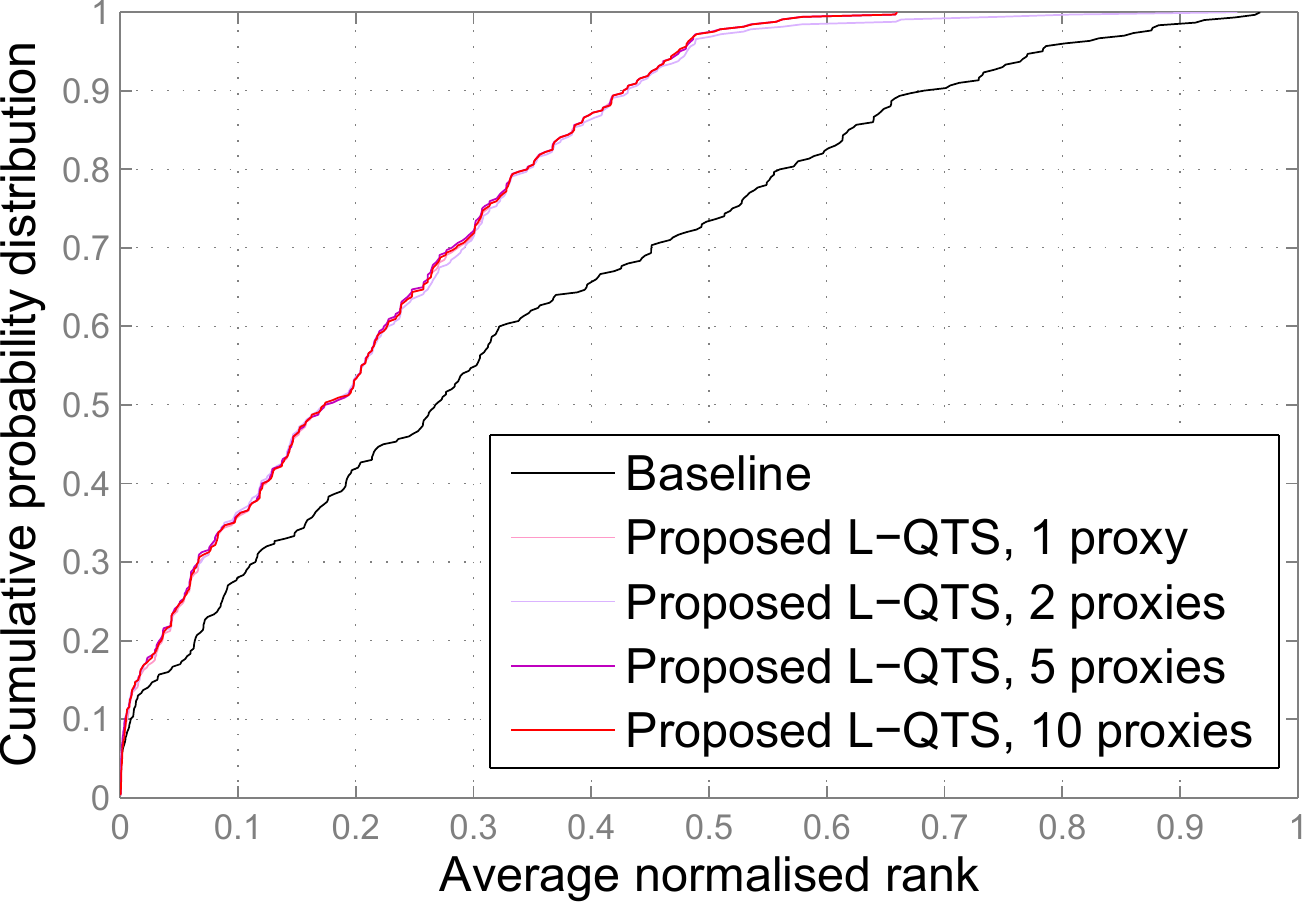}\label{f:anrCCANN}}
  \caption{CDF of the average normalized rank obtained using the exemplar-based (a,b) and subspace-based (c,d) methods.
  (a,c) Comparison of the respective baseline approach, the three simple quasi-transitivity estimation methods, and the proposed learnt quasi-transitivity. (b,d) Comparison of the respective baseline approach and the corresponding proposed method for different numbers of proxies. }
  \label{f:anr}
  \vspace{-10pt}
\end{figure*}

\vspace{-2pt}\subsection{Protocol}\label{ss:protocol}
\begin{figure}[htb]
  \vspace{-0pt}
  \centering
  \subfigure[Inter-class transitivity features]{\includegraphics[width=0.235\textwidth]{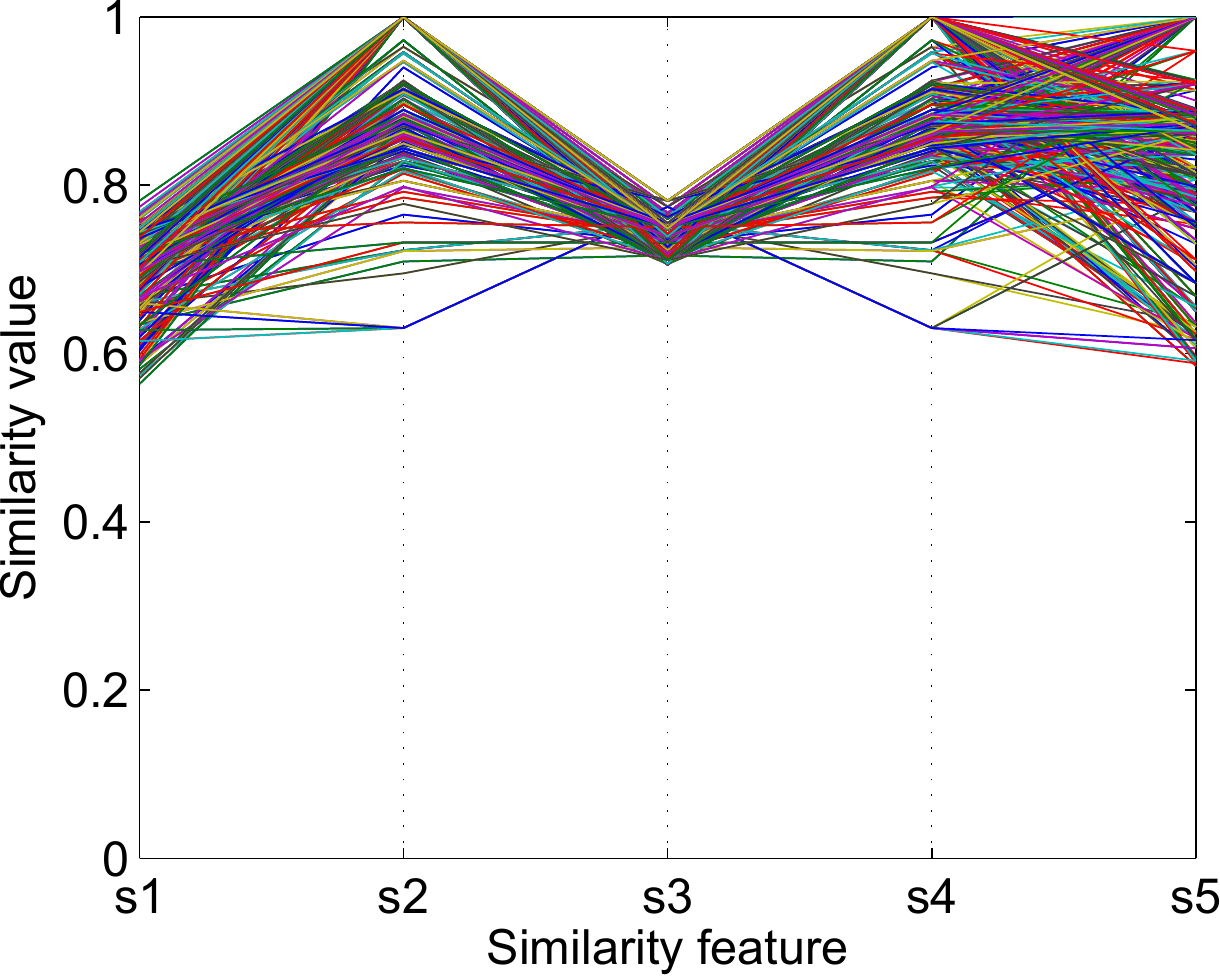}}~~~
  \subfigure[Intra-class transitivity features]{\includegraphics[width=0.235\textwidth]{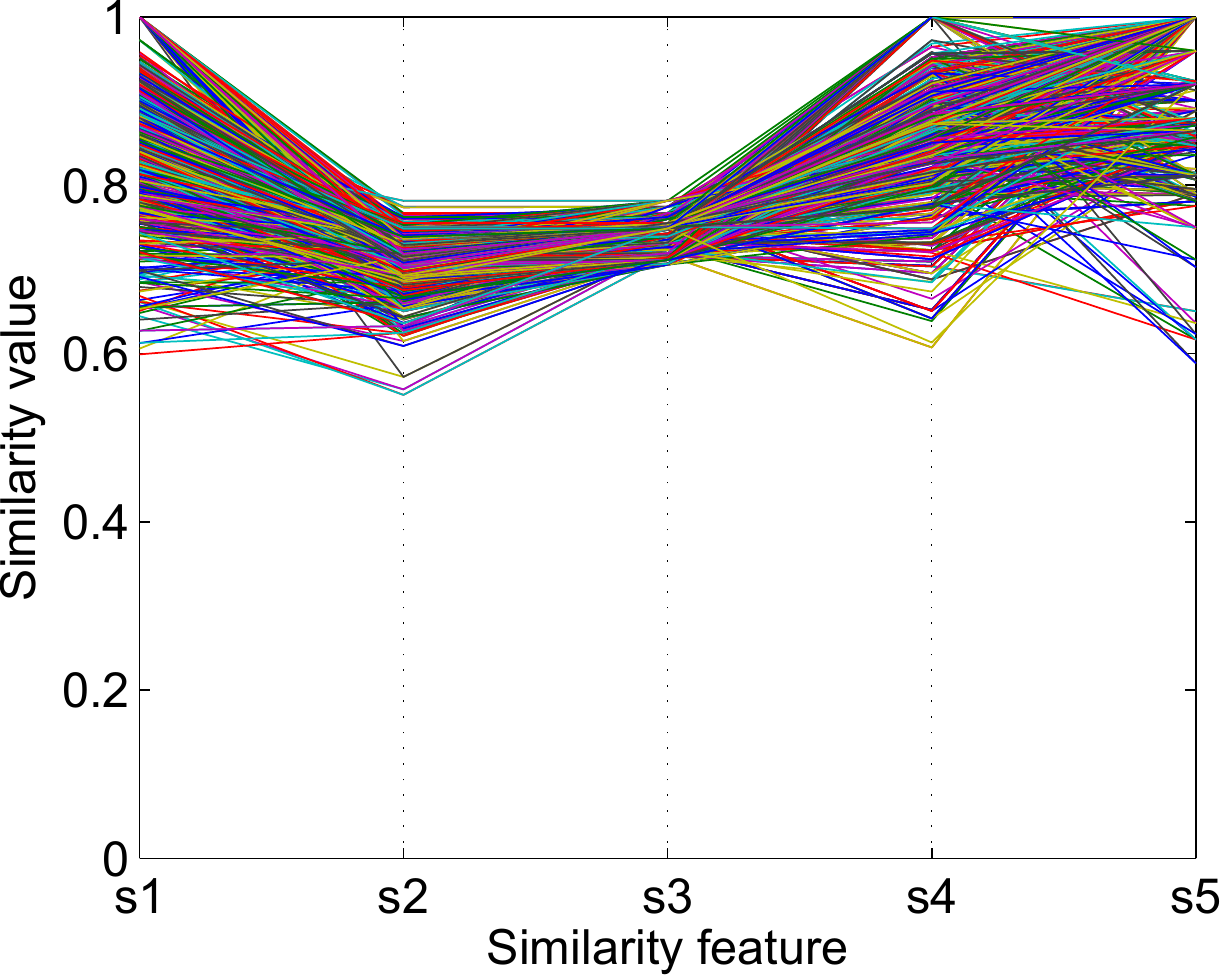}}\\[-8pt]
  \subfigure[Inter-class transitivity features]{\includegraphics[width=0.235\textwidth]{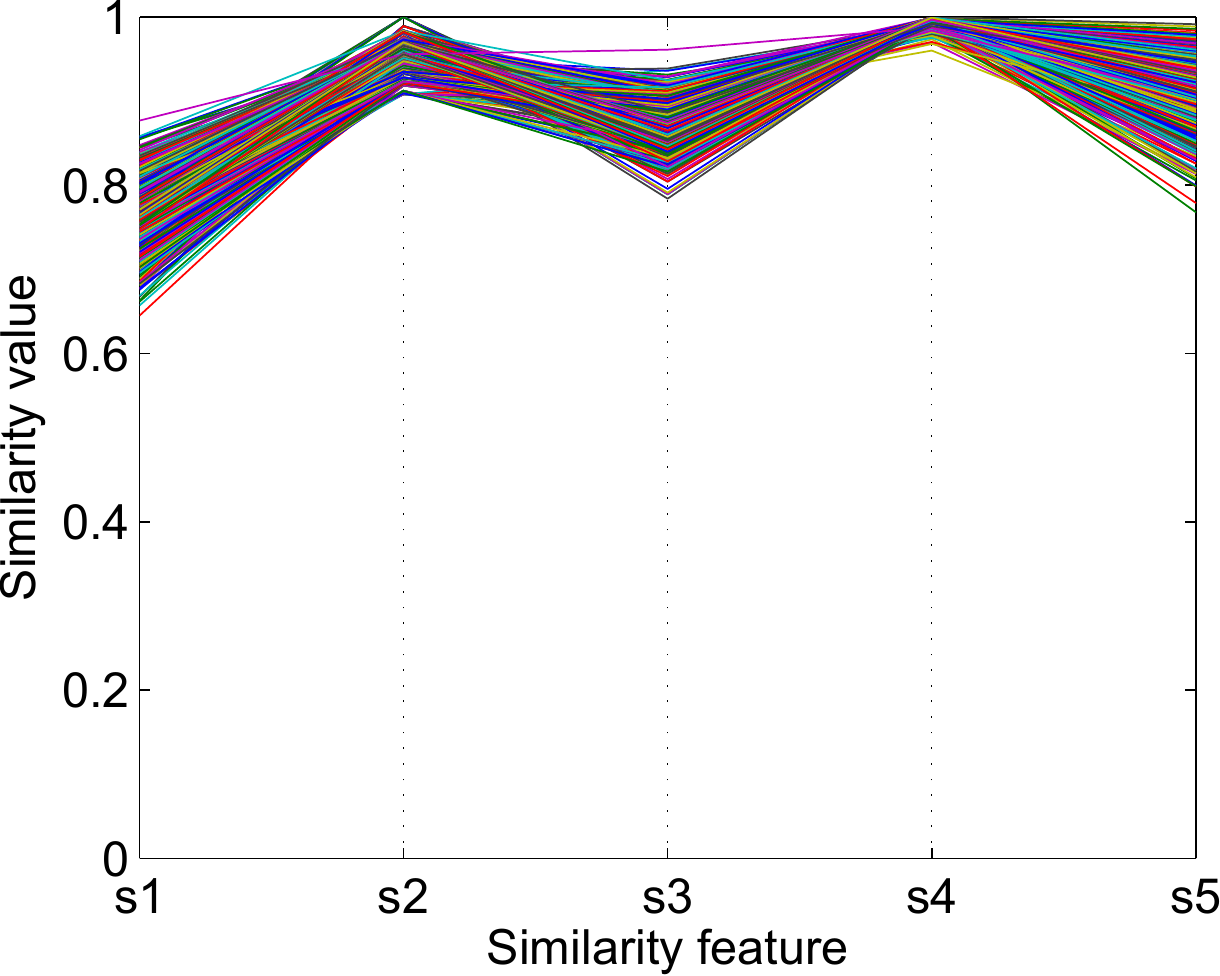}}~~~
  \subfigure[Intra-class transitivity features]{\includegraphics[width=0.235\textwidth]{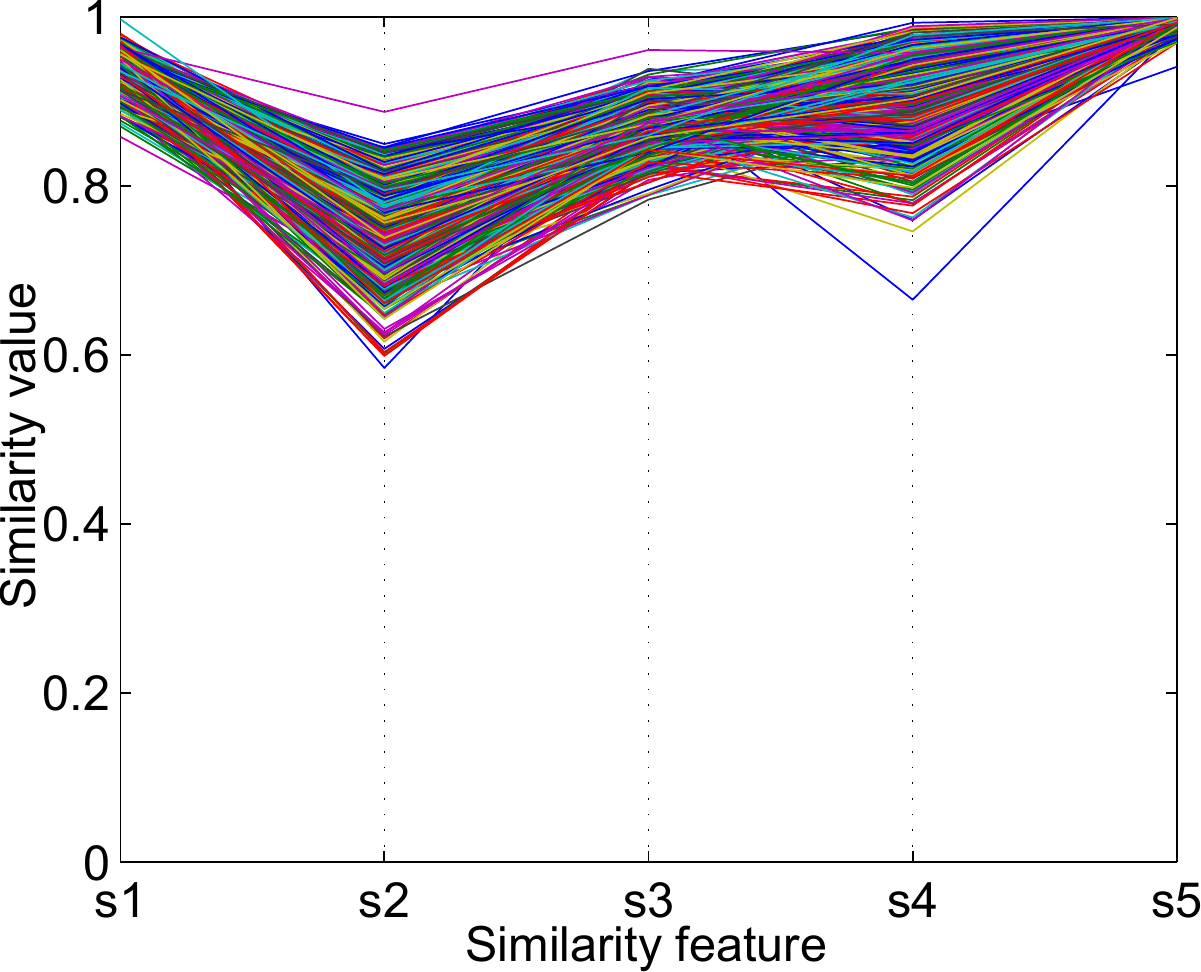}}
  \caption{Training data for the exemplar-based (a,b) and subspace-based (c, d)
  experiments, in the form of intra-class and inter-class transitivity features shown using parallel coordinates. }
  \label{f:features}
  \vspace{-15pt}
\end{figure}

We train the $\epsilon$-SV regressor using 200 randomly selected sets and their proxies (which are not necessarily in the random 200). In principle there is no reason why the entire database would not be used (recall that no labelling or manual intervention is used whatsoever) but we found that 200 sets were sufficient to gather sufficient training data. Examples are shown in Fig~\ref{f:features}; clear patterns are observable both within positive and negative training sets which differ one from another significantly.

The evaluation of the methods described in the previous section was performed by examining all possible retrievals. In other words, we used every set in our database as the query in turn and evaluated the resulting retrieval. To make this feasible we propose a robust sample selection method so as to reduce the computational demands of the otherwise computationally intensive exemplar-based baseline.

\vspace{-15pt}\paragraph{Exemplar baseline: robust sample selection}
It is well established by the existing work on face recognition that the appearance of a face is constrained and thus confined to a region of the image space. Within this region, which is nonlinear, the appearance variation is mostly approximately smooth -- this is sometimes somewhat loosely stated as the face appearance being constrained to a nonlinear appearance manifold~\cite{LuiBeve2008,WangShanChenGao2008}. That being said, the range of appearance variation of a person's face within a \emph{single} video typically covers only a portion of the entirety of possible variation. It is a simple yet important observation that even within this range of appearance the underlying manifold is not uniformly sampled, e.g.\ a person may spend more time in a specific pose than in others. One consequence is that while largely redundant face exemplars of the densely sampled portions of the manifold add little new information about the appearance of the person's face, they can dramatically increase the computational cost of set-based comparisons. This is the case for example for face set-based comparisons which utilize all sample pairs comparisons such as those based on the \textit{maximum maximorum} similarity (i.e.\ all pairs maximum similarity)~\cite{CourSappNaglTask2010} or the \textit{maximum minimorum} distance (a variation of the Hausdorff distance~\cite{ViveSudh2007}). More worryingly, if a sample voting scheme is used~\cite{WolfHassMaoz2011}, redundant exemplars can unduly affect the result even though they carry little additional information.

We overcome both of the problems described above by employing a robust sample selection scheme. Our starting point is the observation that although the intrinsic dimensionality of the entire face manifold is estimated to be in the range 15--22~\cite{Lewi2004}, the appearance variation exhibited in a typical video clip is typically dominated by a single factor such as face yaw changes; the plot in Fig~\ref{f:kpcaEnergy} corroborates this. Led by this insight we employ kernel principal component analysis (KPCA)~\cite{SchoSmolMull1999} to project the original face exemplars onto their dominant nonlinear principal component, uniformly sample the resulting 1D space between the two projections of the two
most extreme exemplars, and finally project them back into the original space. The process is illustrated in Fig~\ref{f:kpca}. The plot in Fig~\ref{f:samplingError} demonstrates that the proposed sample selection does not greatly affect inter-set similarities; a computational improvement of over 2.5 orders of magnitude
(approximately 330 times) was achieved.

\begin{SCfigure*}
  \centering
  \vspace{-15pt}
  \footnotesize
  \begin{tabular}{cc}
    \includegraphics[width =0.35\textwidth]{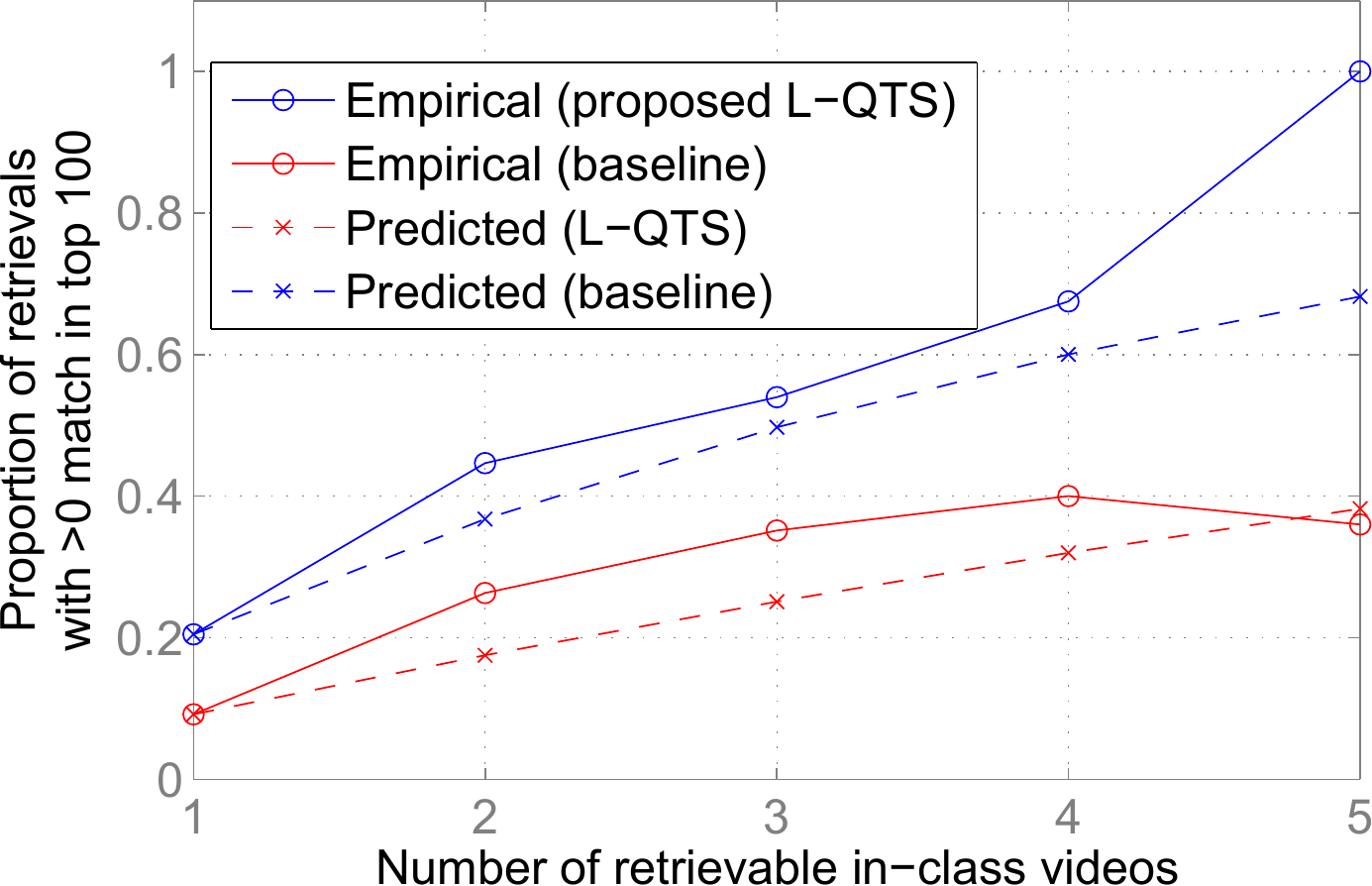}&
  \includegraphics[width=0.35\textwidth]{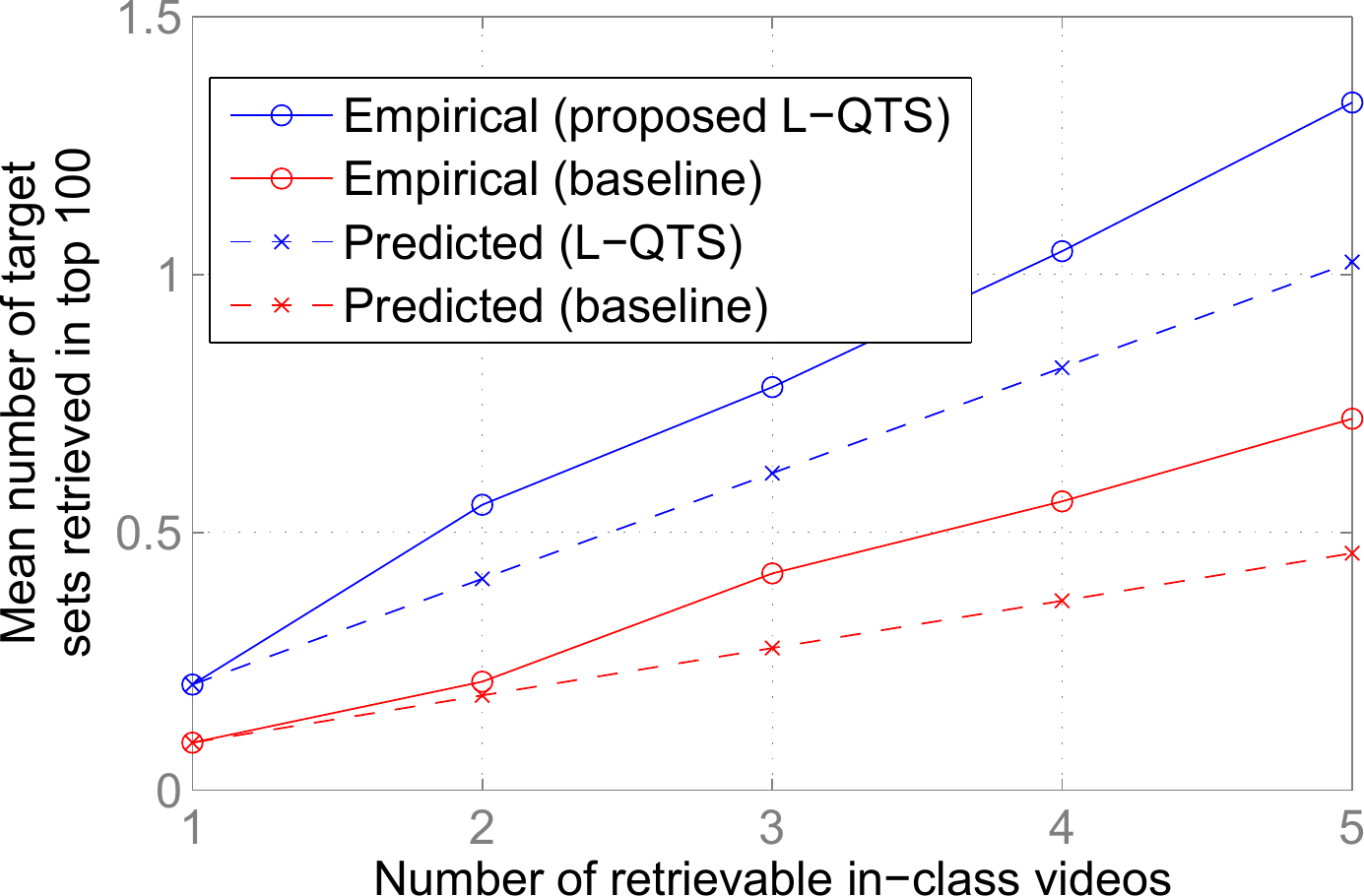}\\
  Match prob.\ w/in rank-100 & Match num.\ w/in rank-100\\
  \end{tabular}
  \caption{Rank-100: (a) probability of a correct match being retrieved, and (b) number of correct matches retrieved, vs.\ number of matches in the database. }
  \label{f:res100}
  \vspace{-0pt}
\end{SCfigure*}

\vspace{-2pt}\subsection{Results and discussion}\label{ss:res}\vspace{-2pt}
The main set of results of our experiments is summarized in the plots in Fig~\ref{f:anrSamp} and~\ref{f:anrCCA} which show the cumulative densities of the ANR achieved for the two baseline methods and different quasi-transitivity approaches. Firstly note that the two baseline methods performed approximately equally well, which is consistent with the previous reports in the literature~\cite{WolfHassMaoz2011}. The three simple attempts at exploiting quasi-transitivity worsened performance significantly, save for the arithmetic mean-based similarity combination for the subspace-based baseline which effected neither an improvement nor deterioration. This
confirmed our expectation expressed in Sec~\ref{ss:features} that the use of inter-personal similarities only is unlikely to be successful and that a richer set of similarity features is needed instead. This leads us to the proposed method which in both cases effected a major performance improvement over both of the baselines. For example, while the exemplar-based baseline produced retrievals with the ANR less than 0.3 in 54.0\% of the cases, the corresponding learnt quasi-transitivity did so in 72.5\% of the cases (an improvement of 34\%). Similarly, while the subspace-based baseline produced retrievals with the ANR less than 0.3 in 54.9\% of the cases, the corresponding learnt quasi-transitivity did so in 72.8\% of the cases. It is particularly interesting to observe in how few cases our method produced bad results (i.e.\ high ANR) -- for both baselines our method achieved ANR lower than 0.5 for over 98\% of retrievals. In contrast, the 98\% quantile of the baseline methods corresponds to the ANR values of 0.92 and 0.88 for the exemplar and subspace-based methods.

The effect of the number of proxies is summarized in Figs~\ref{f:anrSampNN} and~\ref{f:anrCCANN}. For both baselines performance improvement is immediately apparent even using a single proxy per set. Interestingly, while in the case of the exemplar baseline the performance gradually improves up until $k_p=5$, staying approximately steady thereafter, the improvement using the subspace-based baseline is much more dramatic and reaches its peak (on par with the peak of the exemplar baseline) for $k_p=1$ already (ANR plots for different $k_p$ are virtually indistinguishable). Although we are not sure of the exact mechanism that explains this behaviour, it does appear to be linked to the inherent properties of the subspace-based baseline which is additionally supported by the observation that the within-class variability of the corresponding training meta-features is significantly smaller than for the exemplar-based baseline; see Fig~\ref{f:features}.

Let us next turn our attention to the plot in Fig~\ref{f:res100}(a). It shows the proportion of retrievals which result in at least one correct match being retrieved in the top 100 ranked sets as a function of the total number of target sets in the database which correctly match the query. Plotted as solid blue and red lines are the results obtained using the proposed method (with 10 neighbours used as quasi-transitivity proxies) atop of the exemplar-based baseline, and the baseline itself (as expected from Fig~\ref{f:anr}, the results for the subspace-based method are similar and are
thus not included to avoid unnecessary repetition). The plots also show predictions based on the methods' performances for queries in which only a
single correct match is present in the entire database. Specifically, starting from the estimate of the probability $p_{1,100}$ of a correct match being retrieved in the top 100 ranked sets using queries where only a single correct match is possible, if different correct matches are ranked independently when $k$ correct matches exist, the probability of at least
a single correct match being retrieved in the top 100 is approximately $1-(1-p_{1,100})^k$. Since the greatest number of admissible queries (591 individuals in the database have only a single set; these were not meaningful queries for performance evaluation), approximately 48\%, has $k=1$ this is a reasonable estimate to base the prediction on.

Fig~\ref{f:res100}(a) reveals interesting insight into the performance of the proposed method. Specifically, note that unlike the empirical plot of the baseline, the empirical plot of the proposed method grows faster with the number of retrievable sets than the corresponding prediction. This means that the independence assumption underlying the prediction does not hold well, supporting the premise that quasi-transitivity of similarity can be used to improve the retrieval of sets poorly retrieved by the baseline by propagating information from similarly looking individuals or sets of the same person which are acquired in less challenging conditions.

Lastly Fig~\ref{f:res100}(b) shows the average number of correct matches
retrieved in the top 100 ranked sets as a function of the total number of target sets in the database which correctly
match the query. As before the plots also show the corresponding predictions based on the methods' performances for queries in which only a single correct
match is present in the entire database. Starting from  $n_{1,100}$ the average number of correct matches retrieved in the top 100 ranked sets using queries where only a single correct match is possible, if different correct matches are ranked independently when $k$ correct matches exist, the expected number of correct matches in the top 100 is approximately $k\times n_{1,100}$. The improvement effected by the proposed method is again consistent and significant.

\vspace{-2pt}\section{Summary and conclusions}
We introduced a novel framework for improving the performance of retrieval algorithms on large and highly heterogeneous face sets acquired in uncontrolled conditions. In sharp contrast to the previous work, the proposed method learns to benefit from inter-personal \emph{similarity} using what we term quasi-transitivity. A principled and carefully engineered framework performs learning automatically, with no human intervention whatsoever, making our approach readily employable on large data. Effectiveness was demonstrated on the notoriously challenging YouTube database.
\clearpage
\balance

{\small
\bibliographystyle{ieee}
\bibliography{../../../my_bibliography}
}

\end{document}